\pdfoutput=1

\documentclass[11pt]{article}

\usepackage[final]{acl}

\usepackage{times}
\usepackage{latexsym}

\usepackage[T1]{fontenc}

\usepackage[utf8]{inputenc}

\usepackage{microtype}

\usepackage{inconsolata}

\usepackage{graphicx}

\usepackage{amsmath} 
\usepackage{amssymb} 
\usepackage{adjustbox}
\usepackage{booktabs}
\usepackage{array}
\usepackage{subcaption}
\usepackage{multirow}

\title{DSMoE: Matrix-Partitioned Experts with Dynamic Routing for Computation-Efficient Dense LLMs}

\author{
    Minxuan Lv \textsuperscript{\rm 1,2}\footnotemark[1], Zhenpeng Su\textsuperscript{\rm 1,2,3}\footnotemark[1], Leiyu Pan\textsuperscript{\rm 4},Yizhe Xiong\textsuperscript{\rm 5},Zijia Lin\textsuperscript{\rm 3,5}\footnotemark[2],Hui Chen\textsuperscript{\rm 5}\footnotemark[2],Wei Zhou\textsuperscript{\rm 1,2}\\ \textbf{Jungong Han}\textsuperscript{\rm 5},\textbf{Guiguang Ding}\textsuperscript{\rm 5},\textbf{Cheng Luo}\textsuperscript{\rm 3},\textbf{Di Zhang}\textsuperscript{\rm 3},\textbf{Kun Gai}\textsuperscript{\rm 3},\textbf{Songlin Hu}\textsuperscript{\rm 1,2}\footnotemark[2] \\
    \textsuperscript{\rm 1}Institute of Information Engineering, Chinese Academy of Sciences\\
    \textsuperscript{\rm 2}University of Chinese Academy of Sciences \\
    \textsuperscript{\rm 3}Kuaishou Technology,
    \textsuperscript{\rm 4}Tianjin University,
    \textsuperscript{\rm 5}Tsinghua University \\
    \texttt{\{lvminxuan,husonglin\}@iie.ac.cn}  \quad \texttt{huichen@tsinghua.edu.cn} \\
    \texttt{\{suzhenpeng,linzijia\}@kuaishou.com}\\ 
}

\begin{document}
\maketitle
\renewcommand{\thefootnote}{\fnsymbol{footnote}} 
\footnotetext[1]{These authors contributed equally to this work.} 
\footnotetext[2]{Corresponding authors. \\ This work was supported by National Natural Science Foundation of China (Nos.~62271281, 62441235, 62525103). It was also sponsored by CCF-Kuaishou Large Model Explorer Fund (No.~CCF-Kuaishou~2024001).} 
\renewcommand{\thefootnote}{\arabic{footnote}}

\begin{abstract}
As large language models continue to scale, computational costs and resource consumption have emerged as significant challenges. While existing sparsification methods like pruning reduce computational overhead, they risk losing model knowledge through parameter removal. This paper proposes DSMoE (\textbf{D}ynamic \textbf{S}parse \textbf{M}ixture-\textbf{o}f-\textbf{E}xperts), a novel approach that achieves sparsification by partitioning pre-trained FFN layers into computational blocks. We implement adaptive expert routing using sigmoid activation and straight-through estimators, enabling tokens to flexibly access different aspects of model knowledge based on input complexity. Additionally, we introduce a sparsity loss term to balance performance and computational efficiency. Extensive experiments on LLaMA models demonstrate that under equivalent computational constraints, DSMoE achieves superior performance compared to existing pruning and MoE approaches across language modeling and downstream tasks, particularly excelling in generation tasks. Analysis reveals that DSMoE learns distinctive layerwise activation patterns, providing new insights for future MoE architecture design.
\end{abstract}

\begin{figure*}[htbp]
\centering
\includegraphics[width=1\linewidth]{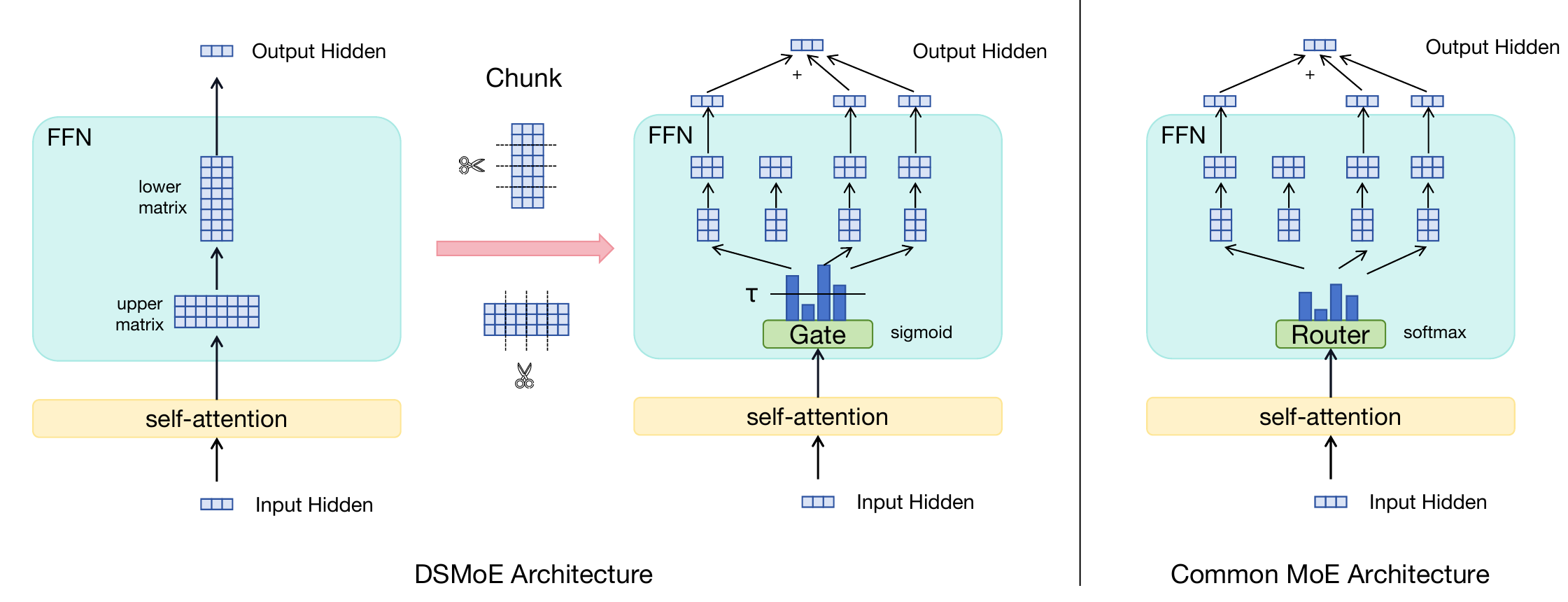}
\caption{The Overview of DSMoE versus Traditional MoE Framework Architectures. The structure shown in the figure is a simplified representation of the transformer backbone. We have simplified the FFN layer structure here; the FFN layer also includes a gating matrix with dimensions matching the upper matrix, which performs Hadamard multiplication with the upper matrix without affecting our partitioning scheme. In the FFN layer, we partition matrices along the intermediate dimension, where portions corresponding to the original matrix multiplication form new expert FFN layers. }
\label{fig:overview}
\end{figure*}

\section{Introduction} 
Large Language Models(LLM) have demonstrated remarkable performance across various downstream tasks\cite{touvron2023llama,dai2022stablemoe,anil2023gemini,biderman2023pythia}. However, as model sizes continue to expand, computational costs and resource consumption grow exponentially. How to improve computational efficiency while maintaining model performance has become a pressing challenge\cite{cheng2024survey}.

At the algorithmic level, approaches to model efficiency optimization generally follow two paradigms: post-training compression and acceleration of dense models, or training of Mixture of Experts (MoE) architectures. While compression methods like pruning achieve efficiency through permanent parameter removal\cite{ashkboos2024slicegpt,ma2023llm,frantar2023sparsegpt}, they may discard valuable knowledge and lack flexibility in handling inputs of varying complexity. Conversely, though MoE approaches effectively expand model capacity\cite{fedus2022switch,dai2024deepseekmoe,liu2024deepseek}, traditional MoE typically employs fixed activation patterns where each token can only access a predetermined number of experts, lacking the ability to dynamically adjust computation based on input complexity. Given that the most widely used and effective foundation models still maintain dense architectures (such as LLaMA\cite{touvron2023llama}, Qwen\cite{bai2023qwen}), we face a critical challenge: how to achieve truly input-adaptive computation while preserving pre-trained knowledge, allowing models to dynamically adjust activated parameters according to varying input complexity, thereby reaching an optimal balance between computational efficiency and model performance.

To address this challenge, we propose DSMoE, a novel approach that partitions pre-trained FFN layers into computational blocks and introduces dynamic routing mechanisms. DSMoE fundamentally differs from existing methods by preserving the original model parameters and reorganizing them into expert networks, while incorporating adaptive routing mechanisms that enable dynamic expert activation based on input complexity, rather than fixed activation strategies. Through straight-through estimator and sparsity loss design, DSMoE enables the model to autonomously learn sparse expert activation patterns, achieving computational resource allocation for inputs of varying complexity.

Extensive experiments conducted on LLaMA-1B and LLaMA-7B models demonstrate encouraging results. Under equivalent computational constraints, our method achieves significant improvements in language modeling perplexity and downstream task performance compared to existing pruning and MoE approaches. Notably superior performance is observed in reasoning and question-answering tasks, particularly in generation tasks.

The main contributions of this work include:

\begin{itemize}
    \item proposing a novel approach that enables transition from dense to dynamically sparse models by preserving and partitioning pre-trained knowledge, enabling different tokens to adaptively access varying portions of model knowledge.
    \item validating the method's effectiveness across multiple benchmarks through extensive experimentation, providing new insights for MoE large model optimization.
\end{itemize}

\section{Related Work}
Model pruning is an effective approach to achieving sparse LLMs while maintaining model functionality. Pruning methods can be categorized into two main types: unstructured and structured pruning. Unstructured pruning operates at the weight level, allowing for arbitrary weight removal \cite{lee2018snip}. In large language models, pruned weights are set to zero \cite{frantar2023sparsegpt,sun2023simple}. However, this method requires specialized hardware and software support for acceleration\cite{han2015deep,wen2016learning,filters2016pruning,tang2021manifold}. Structured pruning takes a coarser-grained approach by removing complete structural units such as convolution kernels, channels, attention heads, or entire layers \cite{you2019gate,ashkboos2024slicegpt,liu2021group,ma2023llm,men2403shortgpt}. Its main advantage is the ability to directly produce regular, narrow model architectures that can achieve acceleration without specialized sparse computation libraries \cite{luo2017thinet,liu2021group,filters2016pruning,nonnenmacher2021sosp}. However, both approaches face a fundamental limitation: achieving efficiency through permanent parameter removal may discard valuable knowledge and lose the ability to adapt computation based on input complexity.

The Mixture of Experts architecture is recognized as a promising approach for model sparsification. Recently, it has garnered significant research attention, with several studies investigating methodologies for converting pre-trained models into MoE architectures. 

While various MoE approaches exist with different objectives, methods like Llama-MoE v2 focus on post-training optimization of instruction-tuned models, and approaches like DTSI target parameter efficiency during training from scratch. However, these methods either address specialized post-training scenarios or require training models from initialization, whereas our approach specifically targets sparsification of pre-trained dense models during the pre-training stage.

MoEfication\cite{zhang2022moefication} trains routers to predict the activation patterns of experts that are partitioned from FFNs while keeping model parameters frozen, thereby activating a fixed number of experts. However, this method was primarily designed for ReLU activation functions and requires additional transformation steps for SiLU/GeLU activation functions that are widely utilized in contemporary Transformer architectures.
FactorLLM\cite{zhao2024factorllm} employs a multi-stage training strategy, initially utilizing the original dense model to guide router training, followed by fixing the router and subsequently training the experts. This sequential training methodology constrains collaborative optimization between routers and experts, and its dependence on a teacher-student framework introduces additional training complexity.
LLaMA-MoE\cite{zhu2024llama} explores the decomposition of FFNs and organizes training according to the Switch Transformer\cite{fedus2022switch} paradigm; however, it merely provides improved expert initialization while lacking flexible input-adaptive computation mechanisms. Given that MoEfication and FactorLLM differ significantly from mainstream MoE methods in architecture design and training paradigms, we choose to use LLaMA-MoE as a comparative approach.

Recent dynamic pruning methods such as DejaVu\cite{liu2023deja} and PowerInfer\cite{song2024powerinfer} can adaptively select activated weights based on input patterns. However, these approaches primarily focus on system-level acceleration through specialized hardware configurations: DejaVu requires integration with asynchronous hardware-aware implementations including kernel fusion and memory coalescing, while PowerInfer employs GPU-CPU hybrid inference engines to exploit locality patterns and minimize communication overhead. In contrast, our method employs algorithm-level sparsification, which reduces the model's floating-point operations.

\section{Background}
For simplicity, we focus on the prevalent architecture of generative large language models while maintaining a concise mathematical formulation. In autoregressive generation tasks, given a sequence $X = (x_1, x_2, ..., x_T)$ of length $T$, the model iteratively produces a probability distribution over the vocabulary for each position conditioned on preceding tokens. This process can be formulated as:
\begin{equation}
\begin{split}
P_{\cdot,t} = \text{softmax}(EH^L_{\cdot,t}) \quad \quad \quad \quad \quad \; \; \, \\
H^L = \text{Transformer}(x_1, x_2, ..., x_{T-1})
\end{split}
\end{equation}

Here, $L$ denotes the number of layers in the Transformer architecture. For any position $t$, $P_{\cdot,t}$ represents the probability distribution over the vocabulary, derived from the $t$-th column of the hidden state matrix $h^L$. Specifically, $H^L = [h_1^L, h_2^L, ..., h_{T-1}^L]$ contains the hidden representations from the final layer, where $h_t^L$ is the contextual embedding at position $t$. The probability of the ground-truth token $x_{t+1}$ is denoted as $P_{x_{t+1},t}$ in the distribution $P{\cdot,t}$. The transformation from hidden states to probability distributions is achieved through a linear projection matrix $E$, followed by a softmax operation.

In typical scenarios, we employ cross-entropy loss for autoregressive learning, which can be expressed as:
\begin{equation}
\mathcal{L}_{\text{LM}} = -\sum_{t=1}^{T-1} \log P(x_{t+1}|x_{\leq t})
\end{equation}

The Transformer architecture consists of multiple layer-wise submodules, where each layer comprises a self-attention module and a Feed-Forward Network (FFN) module. The simplified mathematical formulation can be expressed as:
\begin{equation}
\hat{h^l_t} = \text{Attn}([h^{l-1}_1,h^{l-1}_2,...,h^{l-1}_t]) 
\end{equation}
\begin{equation}
h^l_t = \text{FFN}(\hat{h^l_t})
\end{equation}

FFN modules typically consist of two matrix transformations with a non-linear activation function. In modern language models, the most prevalent FFN implementation uses SwiGLU activation, which involves three essential matrices: the up-projection matrix $\mathbf{U}_{\text{up}}$, the down-projection matrix $\mathbf{V}_{\text{down}}$, and the gate matrix $\mathbf{W}_{\text{gate}}$. The up-projection matrix transforms the input to a higher dimensional space for richer feature representation, the down-projection matrix compresses the information back to the original dimension, and the gate matrix controls information flow through adaptive feature weighting. The FFN output is computed through the following operation:
\begin{equation}
\label{eq:MLP}
h^l_t=(act(\hat{h^l_t}W_{gate}) \odot (\hat{h^l_t}U_{up})) V_{down}
\end{equation}

In this formulation, $\text{act}(\cdot)$ represents the activation function and $\odot$ denotes Hadamard product.

\section{Method}

Although our method is termed DSMoE, its training approach differs from traditional MoE methods such as Switch Transformer and DeepSeeKMoE \cite{dai2024deepseekmoe}. Our objective is to achieve sparsity through partitioning pre-trained models, where each expert inherits a distinct portion of the original model's knowledge. Our approach is based on the principle that the model should learn to selectively utilize different aspects of pre-trained knowledge based on input complexity, rather than routing tokens among independently trained experts. To implement this insight, we present our method in three modules.

\subsection{FFN Partitioning}
The widespread adoption of MoE architectures inspires our exploration of sparsity in FFN layers, suggesting that different parts of computation can be dynamically activated based on input patterns. Previous work has further revealed that FFN layers essentially operate as key-value memories, where different portions of the layer specialize in detecting and processing distinct input patterns\cite{geva2020transformer}. Building on these insights, we propose to directly partition pre-trained FFN layers. As shown in Equation \ref{eq:MLP}, we partition the matrices $\mathbf{U}$, $\mathbf{V}$, and $\mathbf{W}$ into $n$ groups along the intermediate dimension, where each group can be viewed as an ``expert" that inherits a portion of the original transformation capabilities. When summing all expert outputs, this partitioned form is mathematically equivalent to the original FFN computation:
\begin{equation}
\begin{split}
h^l_t = (act(\hat{h^l_t}\begin{bmatrix}
W_1 & \cdots & W_n
\end{bmatrix}) \odot \\
(\hat{h^l_t}\begin{bmatrix}
U_1 & \cdots & U_n
\end{bmatrix})) \begin{bmatrix}
V_1  \\
\vdots  \\
V_n 
\end{bmatrix} \\
=(act(\hat{h^l_t}W_1) \odot \hat{h^l_t}U_1) V_1 + \cdots \\
+ (act(\hat{h^l_t}W_n) \odot \hat{h^l_t}U_n) V_n
\end{split}
\end{equation}

To enable dynamic expert activation based on input, we employ a gating network that determines which experts should be activated. The expert's output is propagated to the subsequent layer only when the corresponding gating activation value exceeds a certain threshold $\tau$. This can be formulated as:
\begin{equation}
\begin{split}
o_i = (act(\hat{h^l_t}W_i) \odot \hat{h^l_t}U_i) V_i \\
h^l_t=\sum^n_{i=1} o_i * G(\sigma(\hat{h^l_t}\mathbf{Y}_i)) \\
G(x) = \begin{cases} x & \text{if } x > \tau \\ 0 & \text{others }  \end{cases}
\end{split}
\label{eq:G_function}
\end{equation}
where $\mathbf{Y} = [\mathbf{Y}_1,\dots,\mathbf{Y}_n] \in \mathbb{R}^{d \times n}$ represents the parameters of the gating network, and $\sigma(\cdot)$ denotes the sigmoid activation function. 

To maintain consistent output norm regardless of the number of active experts, similar to dropout, we scale $h_t^l$ by the ratio of total expert count $n$ to the number of activated experts. This normalization can be expressed as:
\begin{equation}
h_t^l = \frac{n \cdot h_t^l}{\sum_{i=1}^n \mathbb{I}[\sigma(\hat{h}_t^l\mathbf{Y}_k) > \tau]}
\end{equation} 

\begin{table*}[!t]
\centering
\scalebox{0.77}{
\begin{tabular}
{llccrc}
\toprule
Model & Configuration & Params & Activated Params & PPL ($\downarrow$) \\
\midrule
LLaMA-1B& d=2048, D=8192 & 1.24B & 1.24B & 5.67  \\
LLaMA-7B& d=4096, D=11008 & 6.74B & 6.74B & 3.40 \\
\midrule
\textit{LLaMA-1B} \\
\midrule 
LLM-Pruner-channel & d=1215, D=8192 & 889M & 889M  & 7.51 \\
LLM-Pruner-block & d=2048, D=3896.4  & 735M & 735M  & 7.46 \\
SparseGPT & d=2048, D=8192 & 1.24B & 735M  & 9.82 \\
LLaMA-MoE& d=2048, D=1024 \ \ \ $\times 8$, topK=3  & 1.24B & 736M  & 7.45 \\
DSMoE(ours) & d=2048, D=1024 \ \ \ $\times 8$ & 1.24B & 735M & \textbf{7.41} \\
\midrule
\textit{LLaMA-7B} \\
\midrule
LLM-Pruner-channel & d=2401, \ \ D=11008 & 3.95B & 3.95B  & 4.01 \\
LLM-Pruner-block & d=4096, \ \ D=6256.5  & 3.94B & 3.94B  & 4.01 \\
SparseGPT & d=4096, \ \  D=11008 & 6.74B & 3.93B  & 3.96 \\
LLaMA-MoE& d=4096, \ \  D=1376 \ \ \ $\times 8$, topK=3 & 6.74B & 3.98B  & 4.12  \\
DSMoE(ours) & d=4096, \ \ D=1376 \ \ \ $\times 8$ & 6.74B & 3.93B  & \textbf{3.91} \\
\bottomrule
\end{tabular}
}
\caption{Results of perplexity (PPL) across different language models. The \textbf{bold} values indicate the best-performing method among various acceleration approaches. The Configuration column describes the specific model architecture, where $d$ represents the hidden dimension, D denotes the expansion dimension in FFN layers (for LLM-Pruner-block method, this represents the average value), $\times$ n indicates the use of n parallel FFN layers, and topK specifies the number of activated experts per layer in the MoE architecture. The Params column shows the total number of model parameters, while Activated Params indicates the average number of parameters activated during inference.}
\label{tab:PPL_results}
\end{table*}

\subsection{Straight-Through Estimator}
A key challenge in converting dense models to sparse ones is maintaining the learning capability of all experts. During the forward pass, experts with activation values below the threshold $\tau$ do not participate in computation, as defined by the gating function $G(x)$ in Equation \ref{eq:G_function}. However, this thresholding operation creates a critical problem during backpropagation - experts that are not activated receive zero gradients:
\begin{equation}
\begin{split}
    \frac{\partial h_t^l}{\partial \mathbf{V}_i} = \frac{\partial h_t^l}{\partial \mathbf{W}_i} = \frac{\partial h_t^l}{\partial \mathbf{U}_i} = \\ 
    \frac{\partial h_t^l}{\partial \mathbf{Y}_i} = \mathbf{0},   \text{if } \sigma(\hat{h}_t^l\mathbf{Y}_i) \leq \tau
\end{split}
\end{equation}

This gradient blocking prevents non-activated experts from receiving training signals, leading to a ``dead expert" problem where these experts become permanently inactive. Due to random initialization of the sigmoid gating parameters, experts with initially low activation probabilities below the threshold receive zero gradients and cannot improve through training, creating a Matthew effect where inactive experts remain progressively underutilized. Unlike traditional MoE models that train experts from scratch, our experts inherit pre-trained knowledge that we wish to preserve and adapt. To address this issue, we employ the straight-through estimator technique, which allows gradient flow through non-activated experts while maintaining thresholded activation during the forward pass:
\begin{equation}
S(x) = sg(G(x)) + x - sg(x)
\end{equation}
\begin{equation}
    h_t^l = \sum_{i=1}^n o_i \cdot S(\sigma(\hat{h}_t^l\mathbf{Y}_k))
\label{eq:S_function}
\end{equation}
where the operator ``$sg(\cdot)$" is the ``\texttt{stop gradient}" operator to prevent gradient back propagation. The partial derivatives for experts and their gates below the threshold are as follows. Let:
\begin{equation}
\begin{aligned}
a_i = \text{act}(\hat{h}_t^l\mathbf{W}_i), \quad
a'_i = \text{act}'(\hat{h}_t^l\mathbf{W}_i) \,\\[5pt]
g_i = \sigma(\hat{h}_t^l\mathbf{Y}_i), \quad
u_i=\hat{h}_t^l\mathbf{U}_i \quad\quad\quad\quad
\end{aligned}
\end{equation}

The gradients for expert parameters and their gates can be derived as:
\begin{equation}
\adjustbox{scale=1}{$
\frac{\partial h_t^l}{\partial \mathbf{V}_i} = \begin{cases}
(a_i \odot u_i)^\top \cdot g_i & \text{if } g_i > \tau \\
\mathbf{0} & \text{if } g_i \leq \tau
\end{cases}
$}\quad \quad \quad
\end{equation}
\begin{equation}
\adjustbox{scale=0.85}{$
\frac{\partial h_t^l}{\partial \mathbf{W}_i} = \begin{cases}
(\hat{h}_t^l)^\top \odot a'_i \cdot ((u_i \odot \mathbf{V}_i) \cdot g_i) & \text{if } g_i > \tau \\
\mathbf{0} & \text{if } g_i \leq \tau
\end{cases}
$}
\end{equation}
\begin{equation}
\adjustbox{scale=1}{$
\frac{\partial h_t^l}{\partial \mathbf{U}_i} = \begin{cases}
(\hat{h}_t^l)^\top \cdot (a_i \odot \mathbf{V}_i \cdot g_i) & \text{if } g_i > \tau \\
\mathbf{0} & \text{if } g_i \leq \tau
\end{cases}
$}
\end{equation}
\begin{equation}
\adjustbox{scale=0.85}{$
\frac{\partial h_t^l}{\partial \mathbf{Y}_i} = (\hat{h}_t^l)^\top \cdot (o_i \cdot \sigma'(\hat{h}_t^l\mathbf{Y}_i))
$}
\quad \quad \quad \quad \quad \quad 
\end{equation}

The gradient dynamics show a key property: with the straight-through estimator, experts receive gradients for their gating parameters regardless of activation status. The gradient direction for $\mathbf{Y}_i$ depends on whether the expert's output $o_i$ would reduce the overall loss. This allows experts to adaptively learn when to activate based on their usefulness for specific input patterns.

\subsection{Sparse Loss}
Since our experts inherit from a dense model, the model naturally tends to activate all experts to access complete knowledge. However, this conflicts with our goal of sparse computation. We introduce a sparsity loss term that creates an adversarial effect with expert gate gradients, encouraging the model to learn which knowledge is truly necessary for different inputs. 
\begin{equation}
\mathcal{L} = \mathcal{L}_{\text{LM}} + \mathcal{L}_{\text{sparse}}
\end{equation}
where $\mathcal{L}_{\text{sparse}}$ denotes the sparsity loss term, which we abbreviate as $\mathcal{L}{\text{s}}$ in subsequent equations. 
\begin{equation}
\mathcal{L} = \mathcal{L}_{\text{LM}} + \frac{1}{LN}\sum_{l=1}^L\sum_{n=1}^N\mathcal{L}_s(G(\sigma(\hat{h}^l_t\mathbf{Y}_n)))
\end{equation}

We employ $L1$ norm as the sparsity function $\mathcal{L}_s$. Given that our activation function $\sigma(x) > 0$, our final loss function becomes:
\begin{equation}
\mathcal{L} = \mathcal{L}_{LM} + \frac{1}{LN}\sum_{l=1}^L\sum_{n=1}^N G(\sigma(\hat{h}^l_t\mathbf{Y}_n))
\end{equation}

The gradients introduced by this sparse loss term create an adversarial effect with the gate gradients, encouraging the model to actively suppress the output of less important experts across different layers.

It is worth noting that our approach differs fundamentally from the MoE framework and therefore does not require auxiliary load balancing losses. While load balancing losses in MoE aim to ensure uniform training across experts, our objective is solely focused on learning sparse activation patterns. Furthermore, unlike MoE which typically enforces a fixed number of active experts, our method allows for flexible activation patterns determined by the learned gating mechanism.

\section{Experiments}

\begin{table*}[!t]
\centering
\scalebox{0.70}{
\begin{tabular}{lp{1.40cm}<{\centering}p{1.40cm}<{\centering}p{1.40cm}<{\centering}p{1.30cm}<{\centering}p{1.30cm}<{\centering}p{1.40cm}<{\centering}p{1.30cm}<{\centering}p{1.40cm}<{\centering}p{1.40cm}<{\centering}p{1.40cm}<{\centering}p{1.40cm}<{\centering}}
\toprule
Model & Hellaswag & LAMBADA & PIQA & SIQA & StoryCloze & Wino & GSM8K & NaturalQs & TriviaQA & WebQs \\
\midrule
LLaMA-1B & 64.09 & 61.05 & 75.51 & 42.47 & 72.58 & 60.85 & 4.85 & 12.52 & 36.08 & 22.49 \\
LLaMA-7B & 76.39 & 72.34 & 79.05 & 44.67 & 79.15 & 70.87 & 14.70 & 26.28 & 61.89 & 32.82 \\
\midrule
\textit{LLaMA-1B} \\
\midrule
LLM-Pruner-channel & 53.44 & 45.04 & 71.43 & 40.94 & 68.67 & \textbf{58.45} & 1.44 & 6.98 & 17.46 & 14.56 \\
LLM-Pruner-block & 51.05 & 46.28 & 71.71 & 41.04 & 68.62 & 56.27 & 1.36 & 7.28 & 18.46 & 14.56 \\
SparseGPT & \textbf{54.01} & \textbf{56.49} & 71.10 & 40.68 & 68.05 & 57.30 & 1.51 & 5.29 & 14.44 & 11.61 \\
LLaMA-MoE & 49.06 & 44.84 & 70.02 & 41.05 & 65.47 & 55.64 & 1.62 & 5.76 & 13.49 & 11.27 \\
DSMoE(ours) & 50.92 & 48.12 & \textbf{72.36} & \textbf{41.14} & \textbf{68.78} & 56.35 & \textbf{1.67} & \textbf{8.17} & \textbf{25.52} & \textbf{18.21} \\
\midrule
\textit{LLaMA-7B} \\
\midrule
LLM-Pruner-channel & 66.41 & 61.63 & 74.97 & 43.19 & 75.30 & 66.85 & 4.85 & 12.63 & 36.02 & 20.57 \\
LLM-Pruner-block & 67.93 & 62.02 & 76.22 & 44.26 & 75.46 & 63.53 & 1.81 & 12.96 & 38.77 & 21.65 \\
SparseGPT & \textbf{73.60} & 67.43 & 77.36 & 44.21 & \textbf{76.37} & \textbf{70.48} & \textbf{8.33} & 17.61 & 47.83 & 24.90 \\
LLaMA-MoE & 63.89 & 60.49 & 74.10 & 43.29 & 72.90 & 61.17 & 3.26 & 11.58 & 31.25 & 19.09 \\
DSMoE(ours) & 70.22 & \textbf{67.61} & \textbf{78.12} & \textbf{44.31} & \textbf{76.37} & 66.77 & 6.41 & \textbf{22.04} & \textbf{57.94} & \textbf{29.92} \\
\bottomrule
\end{tabular}
}
\caption{Performances of language models on downstream tasks. The best score is marked in \textbf{bold}.}
\label{tab:benchmark_score}
\end{table*}

\subsection{Dataset}
We gathered datasets from various domains to continually pre-train the base model. For the general domain, we used the Fineweb-edu dataset, which consists of high-quality educational web pages filtered from the Fineweb dataset \cite{penedo2024fineweb}. In the math and coding domains, we selected the OpenWebMath \cite{DBLP:conf/iclr/PasterSAB24} and StarCoder \cite{DBLP:journals/tmlr/LiAZMKMMALCLZZW23} datasets respectively. The OpenWebMath dataset contains high-quality mathematical text data extracted from web pages, while the StarCoder dataset offers a diverse range of code data and has been demonstrated to effectively pre-train well-behaved code models. Furthermore, it has been demonstrated that  incorporating synthetic data enhances model pre-training performance \cite{abdin2024phi}. Therefore, we introduced the Cosmopedia dataset to leverage this advantage\cite{benallal2024cosmopedia}.

Furthermore, we mixed datasets from different domains. Due to computational resource limitations, we set the total amount of training data to 10 billion tokens. 
Finally, we used the tokenizers from LLaMA to segment the data, limiting the maximum sample length to 1024 tokens for each. We randomly sampled 5,000 non-overlapping instances from each dataset as the validation set, ensuring no intersection with the training set.

\subsection{Experimental Setup}
We evaluate DSMoE on two pre-trained models of different scales: Llama-7B\footnote{\url{https://huggingface.co/meta-llama/Llama-2-7b}} and Llama-1B\footnote{\url{https://huggingface.co/meta-llama/Llama-3.2-1B}}. For our method's hyperparameters, we simply set the activation threshold $\tau=0.5$, learning rate to 2e-5, batch size to 32, and sequence length to 1024. To ensure fair evaluation, all baseline methods underwent continued training on identical data quantities (10B tokens) with the same training configurations. 

We compare our approach with several baselines: the channel-wise and block-wise methods from LLM-Pruner (a structured pruning approach), and SparseGPT (an unstructured pruning method). To ensure fair comparison, all baseline methods (LLM-Pruner channel/block-wise, SparseGPT, and LLaMA-MoE) were trained on identical data quantities (10B tokens) and configured to match DSMoE's activated parameter count.

Additionally, we compare against LLaMA-MoE, which applies a similar FFN partitioning scheme but follows the traditional MoE paradigm with fixed top-k expert selection and standard MoE training objectives, to investigate whether conventional MoE frameworks can effectively leverage pre-trained weights through warm-starting.

\subsection{Main Results}

We first present the model's perplexity on the validation set. 
Following previous work\cite{touvron2023llama,brown2020language,su2024cartesianmoe,xiong2024temporal,dai2024deepseekmoe}, we then evaluate the model's performance on downstream benchmarks, which includes zero-shot accuracy testing on HellaSwag\cite{zellers2019hellaswag}, LAMBADA\cite{paperno2016lambada}, SIQA\cite{sap2019socialiqa}, PIQA\cite{bisk2020piqa}, StoryCloze\cite{mostafazadeh2016corpus}, and Winogrande\cite{sakaguchi2021winogrande}. Additionally, we conduct 5-shot evaluation measuring exact match performance on TriviaQA\cite{joshi2017triviaqa}, WebQuestions (WebQs)\cite{berant2013semantic}, GSM8K\cite{cobbe2021training}, and Natural Questions (NaturalQs)\cite{kwiatkowski2019natural}.

\subsubsection{Perplexity Results}
Table \ref{tab:PPL_results} presents the perplexity results of the baseline dense model and its pruned, sparsified variants. The results demonstrate that DSMoE consistently outperforms baseline models under equivalent activation constraints. Our experimental results indicate that DSMoE achieves superior efficiency compared to static parameter pruning. Furthermore, DSMoE exhibits better performance than fixed-activation methods like MoE, which can be attributed to the fact that knowledge from all experts contributes to the model's learning process, enabling it to develop the ability to flexibly select activations based on input. Additionally, DSMoE exhibits distinctive feature processing capabilities, learning layer-specific activation patterns that naturally emerge from the input complexity. We will examine these emergent patterns in detail in the analysis section.

In conclusion, DSMoE demonstrates consistent superiority across models of two different scales, highlighting its robust advantages.

\subsubsection{Benchmark Results}


Table \ref{tab:benchmark_score} presents the benchmark performance of various pruning methods, traditional MoE approaches, and DSMoE. DSMoE achieved the best performance in 7 out of 10 benchmarks for both LLaMA-1B and LLaMA-7B model architectures, demonstrating superior effectiveness over existing sparsification methods across most evaluation metrics.

Specifically, DSMoE exhibited excellent performance on inference tasks (i.e., the first 6 benchmarks), achieving the best results on PIQA, SIQA, and StoryCloze test sets. While not achieving top performance on Hellaswag, LAMBADA, and Wino test sets, DSMoE still ranked among the leading models. For generation tasks (i.e., the last 4 benchmarks), DSMoE demonstrated remarkable effectiveness. Apart from slightly lower performance on GSM8K with LLaMA-7B compared to SparseGPT, it significantly outperformed other sparse methods on all other test sets, with performance only a few points below the dense model. These results highlight DSMoE's potential, particularly in generation tasks.

Furthermore, we observed that the performance gap between DSMoE and other sparse approaches was more pronounced in LLaMA-7B compared to LLaMA-1B. This may be attributed to greater model redundancy at larger parameter scales, enabling DSMoE to more effectively prune unnecessary information. This observation suggests the potential scalability of DSMoE to models with larger parameter counts.

\section{Analyses}
\subsection{Ablation Study: Removing Straight-Through Estimator}
To validate the necessity of the straight-through estimator mechanism in DSMoE, we conduct an ablation study by removing this component. Specifically, instead of using Equation (\ref{eq:S_function}) for training, we employ Equation (\ref{eq:G_function}). We perform this comparative analysis on the LLaMA-1B model.

\begin{table}[!t]
\centering
\scalebox{0.70}{
\begin{tabular}{lcc}
\toprule
Model & DSMoE & \textit{w/o $S(x)$} \\
\midrule
Hellaswag & 50.92 & 32.29 \\
LAMBADA & 48.12 & 27.79 \\
PIQA & 72.36 & 62.73 \\
SIQA & 41.14 & 39.30 \\
StoryCloze & 68.67 & 57.14 \\
Wino & 56.35 & 50.83 \\
GSM8K & 1.67 & 0.38 \\
NaturalQs & 8.17 & 2.47 \\
TriviaQA & 25.52 & 2.95 \\
WebQs & 18.21 & 1.00 \\
\midrule
PPL & 7.41 & 12.75 \\
\bottomrule
\end{tabular}
}
\caption{Ablation study of DMoE against the model without direct estimation function S(x), where G(x) is employed in place of S(x).}
\label{tab:ab_study_1}
\end{table}

As shown in Table \ref{tab:ab_study_1}, the model without straight-through estimator significantly underperforms the complete model in terms of both perplexity and benchmark performance. This substantial degradation occurs because routing parameters for non-activated experts receive zero gradients during backpropagation, preventing these routes from being adjusted to utilize more of the pre-trained knowledge inherited from the dense model. Without the ability to adaptively modify routing decisions, potentially valuable knowledge encoded in these experts becomes permanently inaccessible, leading to significant performance loss.

\subsection{Ablation Study: Training without Piecewise Function G(x)}
To validate the necessity of incorporating piecewise function learning during training, we conduct an ablation study by removing the piecewise function G(x) and using the following formula for training:
\begin{equation}
h^l_t=\sum^n_{i=1} o_i * \sigma(\hat{h^l_t}Y_i)
\end{equation}

\begin{figure}[htbp]
\centering
\includegraphics[width=0.8\linewidth]{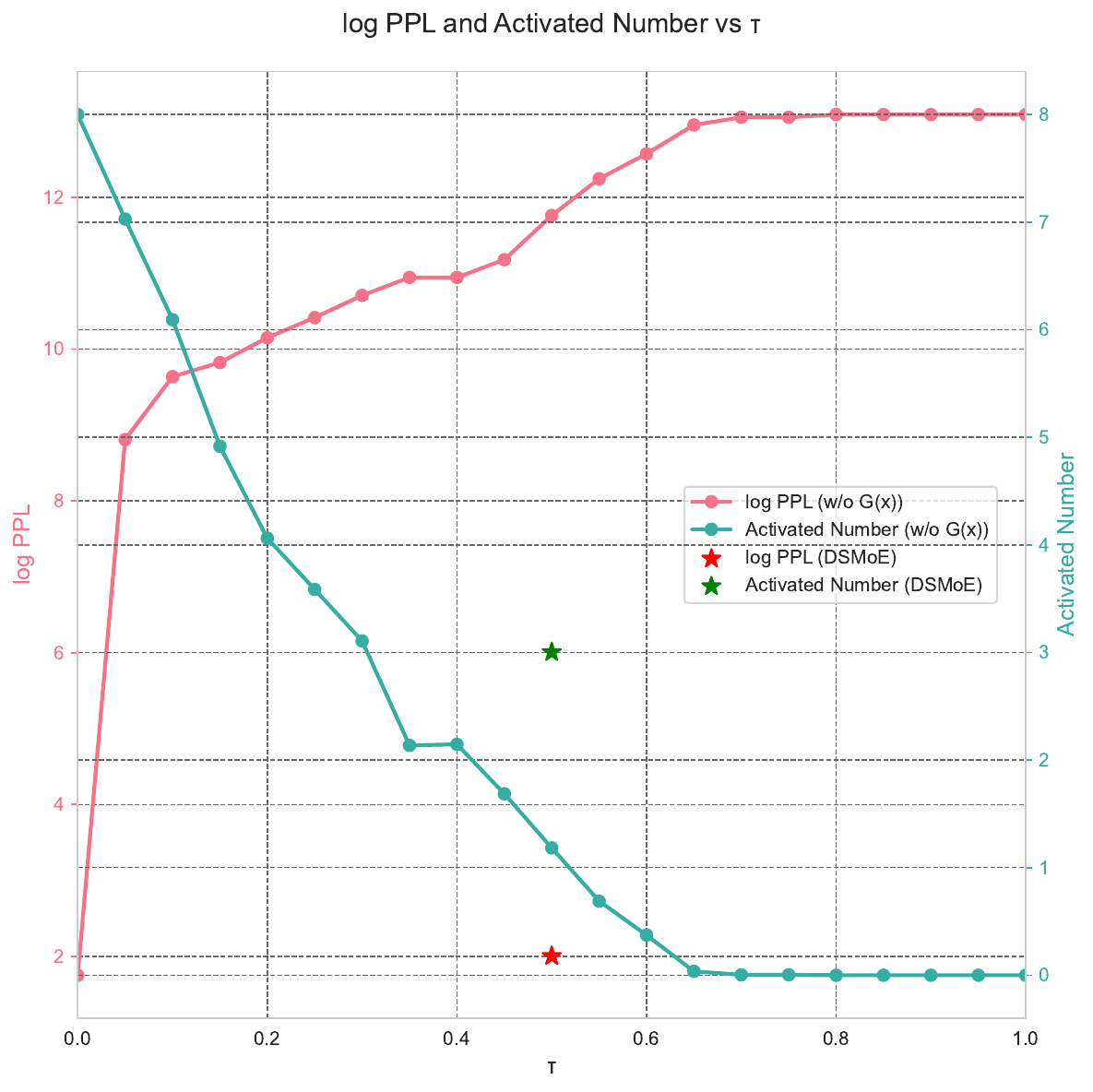}
\caption{During the training phase, G(x) is not utilized. In the inference phase, G(x) is employed for activation. The model's perplexity and the number of activated experts vary with the threshold $\tau$. The pentagram markers indicate the perplexity and number of activated experts achieved by DSMoE.}
\label{fig:diff_threshold}
\end{figure}

Prior to inference, we determine the appropriate activation level by adjusting the threshold value on the validation set, with a step size of 0.05. Figure \ref{fig:diff_threshold} illustrates the relationship between perplexity and the average number of activated experts on the validation set.

The results clearly demonstrate that as the threshold increases, perplexity rises rapidly while the average number of activated experts decreases correspondingly. This observation indicates that without the piecewise function G(x), all experts participate in computation and gradient updates. Under the constraint of sparsity loss, the model tends to distribute activation values uniformly across all experts rather than learning to distinctively identify more important experts. This leads to two consequences: first, the activation values for each expert are suppressed to a relatively low level, and second, the learned importance of each expert becomes relatively uniform. Under the same activation constraints as DSMoE, the approach without the piecewise function G(x) exhibits higher perplexity, highlighting how this training-inference inconsistency significantly degrades model performance.

\subsection{Layer-wise Activation Patterns Analysis}

\begin{figure}[htbp]
    \centering
    \begin{subfigure}{0.47\linewidth}
        \centering
        \includegraphics[width=\linewidth]{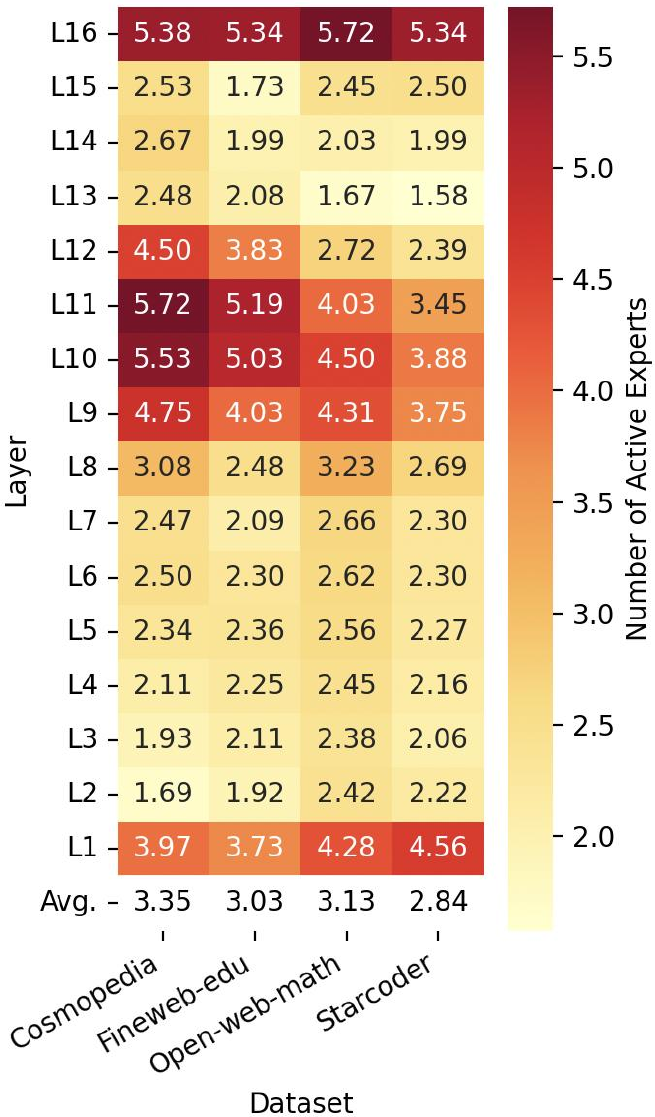}
        \caption{Heatmap for 1B model}
        \label{fig:hotmap1B}
    \end{subfigure}
    \hfill
    \begin{subfigure}{0.51\linewidth}
        \centering
        \includegraphics[width=\linewidth]{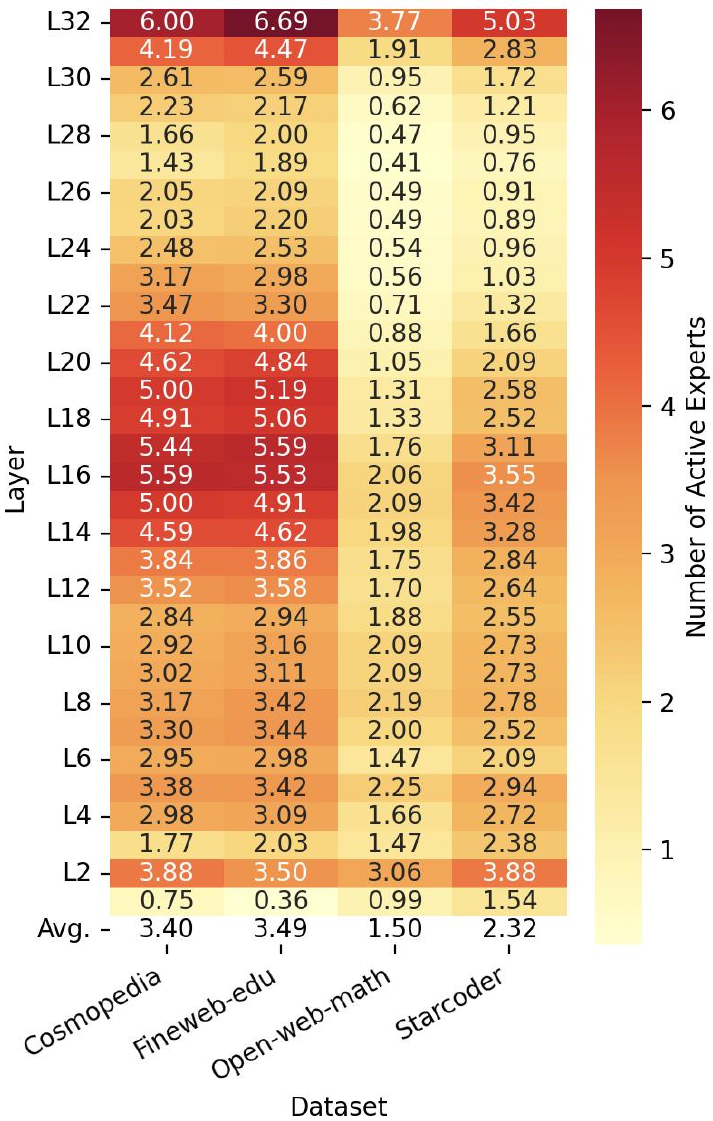}
        \caption{Heatmap for 7B model}
        \label{fig:hotmap7B}
    \end{subfigure}
    \caption{Heatmap visualization of expert activation counts across different layers and average expert activations for LLaMA-7B and LLaMA-1B models on various validation sets.}
    \label{fig:heatmaps}
\end{figure}

We evaluated DSMoE across different validation sets and generated heatmaps to visualize the distribution of activated experts across network layers. Both model sizes exhibit a distinctive activation pattern: higher activation counts at both input and output layers, elevated activation in middle layers, and lower activation in remaining layers - forming a ``W-shaped" pattern.

The bottom layers, which typically encode fundamental features, demonstrate high expert activation. This suggests the model's tendency to activate multiple experts in parallel to process multi-dimensional input features, potentially serving as an ``information preservation mechanism" to retain critical base-level information. The top layers, responsible for final decision-making and output generation, show increased expert activation to enhance output robustness by reducing individual expert bias through collective decision-making. The elevated activation in middle layers suggests these layers serve as critical zones for feature transformation, integration, and processing of long-range dependencies. This bottom-middle-top activation pattern forms a complete information processing pipeline: bottom layers for extensive collection and processing of basic features, middle layers for feature transformation and information integration, and top layers for comprehensive decision-making and output generation.

Furthermore, we observed significant variations in both the average number of activated experts and activation patterns across different test sets. This indicates that DSMoE implements dynamic regulation mechanisms specific to different inputs rather than converging to a homogeneous learning pattern.

These observations provide novel insights for future MoE architectures, suggesting that expert activation counts can be strategically varied across different layers of the network.

\section{Conclusion}

This paper presents DSMoE, a novel approach that achieves model sparsification by partitioning pre-trained FFN layers into computational blocks. Experiments on LLaMA models demonstrate superior performance over existing pruning and MoE approaches under equivalent computational constraints, while revealing distinctive layerwise activation patterns for future MoE designs.

\section{Limitations}
Due to computational resource constraints, we were only able to evaluate DSMoE on language models up to 7B parameters. Future work with access to larger computational resources could explore the scalability and effectiveness of our approach on larger model architectures, which may reveal additional insights about the relationship between model scale and dynamic sparsification patterns.

\bibliography{custom}

\appendix

\section{Sparseness Adjustment}
Although our method does not explicitly specify the activation quantity (sparsity degree) of the MoE model, the sparsity of DSMoE can be adjusted by modulating the hyperparameter $\tau$. The specific regulatory effects are shown in the table \ref{tab:tau}.

\begin{table}[htbp]
\centering
\caption{DSMoE LLaMA Models: Threshold ($\tau$) vs. Performance and Parameter Activation}
\small
\begin{tabular}{c|cc|cc}
\hline
\multirow{2}{*}{$\tau$} & \multicolumn{2}{c|}{LLaMA-7B} & \multicolumn{2}{c}{LLaMA-1B} \\
\cline{2-5}
 & PPL & activated params & PPL & activated params \\
\hline
0.2 & 3.82 & 65.45\% & 7.22 & 64.19\% \\
0.3 & 3.83 & 62.70\% & 7.24 & 62.32\% \\
0.4 & 3.85 & 60.43\% & 7.29 & 60.79\% \\
0.5 & 3.91 & 58.46\% & 7.41 & 59.35\% \\
0.6 & 4.02 & 56.54\% & 7.61 & 57.87\% \\
0.7 & 4.28 & 54.77\% & 8.01 & 56.34\% \\
0.8 & 5.09 & 52.54\% & 8.85 & 54.37\% \\
\hline
\end{tabular}
\label{tab:tau}
\end{table}

The results demonstrate that as $\tau$ increases from 0.2 to 0.8, perplexity gradually increases while the percentage of activated parameters decreases, which aligns with intuitive expectations. Performance degradation is relatively modest in the range of $\tau$=0.2 to $\tau$=0.5, but becomes more pronounced beyond $\tau$=0.5.

We selected $\tau$=0.5 as the default value for our main experiments because it offers an optimal balance between model performance and computational efficiency. In practical applications, $\tau$ can function as an adjustable parameter that users can tune according to their specific computational resource constraints and performance requirements.

\section{Impact of Continued Pretraining Token Count on DSMoE Performance}
To evaluate how the number of tokens used in continued pretraining affects DSMoE performance, we conducted a series of controlled experiments on both LLaMA-7B and LLaMA-1B models. Tables \ref{tab:token_impact_1b} and \ref{tab:token_impact_7b} show the perplexity changes for both models across different token counts. Our approach achieves relatively favorable performance even with fewer tokens, illustrating the relationship between training tokens and complexity (PPL). Performance tends to stabilize after approximately 8 billion training tokens.

\begin{table}[ht]
\centering
\tiny
\begin{tabular}{lcccccc}
\toprule
Tokens (B) & 2.2 & 3.8 & 5.4 & 7.0 & 7.8 & 8.6 \\
\midrule
PPL & 7.384 & 7.323 & 7.481 & 7.488 & 7.445 & 7.422 \\
\bottomrule
\end{tabular}
\caption{Effect of token count on LLaMA-1B DSMoE model performance}
\label{tab:token_impact_1b}
\end{table}

\begin{table}[ht]
\centering
\tiny
\begin{tabular}{lcccccc}
\toprule
Tokens (B) & 2.4 & 3.2 & 4.8 & 6.4 & 8.0 & 9.6 \\
\midrule
PPL & 4.091 & 4.029 & 3.994 & 3.975 & 3.929 & 3.916 \\
\bottomrule
\end{tabular}
\caption{Effect of token count on LLaMA-7B DSMoE model performance}
\label{tab:token_impact_7b}
\end{table}



\end{document}